\documentclass[11pt]{article}
\usepackage{acl2015}
\usepackage{times}
\usepackage{url}
\usepackage{latexsym}
\usepackage{multirow}
\usepackage{amssymb}
\usepackage{amsthm}
\usepackage{color}
\usepackage{amsmath}
\usepackage{verbatim}
\usepackage{tikz-qtree}
\usepackage{tikz}
\usetikzlibrary{shapes.geometric}
\usetikzlibrary{shapes.arrows}
\usetikzlibrary{patterns}
\usetikzlibrary{chains}
\usetikzlibrary{calc}

\newcommand{\ensuretext}[1]{#1}
\newcommand{\ignore}[1]{}
\newcommand{\mycomment}[3]{\ensuretext{\textcolor{#3}{[#1 #2]}}}
\newcommand{\cjdmarker}{\ensuretext{\textcolor{blue}{\ensuremath{^{\textsc{C}}_{\textsc{D}}}}}}
\newcommand{\cjd}[1]{\mycomment{\cjdmarker}{#1}{blue}}
\newcommand{\nascomment}[1]{\textcolor{blue}{\textbf{[#1 --\textsc{nas}]}}}

\newsavebox{\one}
\newsavebox{\two}
\newsavebox{\three}
\newsavebox{\four}
\newsavebox{\five}

\title{\mbox{Transition-Based Dependency Parsing with Stack Long Short-Term Memory}}
\author{Chris Dyer$^{\clubsuit\spadesuit}$ ~ Miguel Ballesteros$^{\diamondsuit\spadesuit}$ ~ Wang Ling$^{\spadesuit}$ ~ Austin Matthews$^\spadesuit$ ~ Noah A. Smith$^{\spadesuit}$\\
  $^\clubsuit$Marianas Labs ~~ $^\diamondsuit$NLP Group, Pompeu Fabra University ~~ $^\spadesuit$Carnegie Mellon University \\
{ \tt chris@marianaslabs.com,  miguel.ballesteros@upf.edu,} \\ { \tt $\{$lingwang,austinma,nasmith\}@cs.cmu.edu }
}
\date{}

\begin{document}
\maketitle
\begin{abstract}
We propose a technique for learning representations of parser states in transition-based dependency parsers.  Our primary innovation is a new control structure for sequence-to-sequence neural networks---the stack LSTM. Like the conventional stack data structures used in transition-based parsing, elements can be pushed to or popped from the top of the stack in constant time, but, in addition, an LSTM maintains a continuous space embedding of the stack contents. This lets us formulate an efficient parsing model that captures three facets of a parser's state: (i) unbounded look-ahead into the buffer of incoming words, (ii) the complete history of actions taken by the parser, and (iii) the complete contents of the stack of partially built tree fragments, including their internal structures. Standard backpropagation techniques are used for training and yield state-of-the-art parsing performance.
\end{abstract}

\section{Introduction}

Transition-based dependency parsing formalizes the parsing problem as a series of decisions that read words sequentially from a buffer and combine them incrementally into syntactic structures \cite{yamada03,nivre03iwpt,nivre2004}. This formalization is  attractive since the number of operations required to build any projective parse tree is linear in the length of the sentence, making transition-based parsing computationally efficient relative to graph- and grammar-based formalisms. The challenge in transition-based parsing is modeling which action should be taken in each of the unboundedly many states encountered as the parser progresses.

This challenge has been addressed by development of alternative transition sets that simplify the modeling problem by making better attachment decisions \cite{nivre07naacl,nivre08cl,nivre09acl,ChoiM13,bohnet-nivre:2012:EMNLP-CoNLL}, through feature engineering \cite{zhang-nivre:2011:ACL-HLT2011,BallesterosNivre2014,chen-zhang-zhang:2014:Coling,ballesteros2014automatic} and more recently using neural networks \cite{chen:2014,stenetorp:2013}.

We extend this last line of work by learning representations of the parser state that are sensitive to the complete contents of the parser's state: that is, the complete input buffer, the complete history of parser actions, and the complete contents of the stack of partially constructed syntactic structures. This ``global'' sensitivity to the state contrasts with previous work in transition-based dependency parsing that uses only a narrow view of the parsing state when constructing representations (e.g., just the next few incoming words, the head words of the top few positions in the stack, etc.). 
Although our parser integrates large amounts of information, the representation used for prediction at each time step is constructed incrementally, and therefore parsing and training time remain linear in the length of the input sentence. The technical innovation that lets us do this is a variation of recurrent neural networks with long short-term memory units (LSTMs) which we call \textbf{stack LSTMs} (\S\ref{sec:stacklstms}), and which support both reading (pushing) and ``forgetting'' (popping) inputs. 

Our parsing model uses three stack LSTMs: one representing the input, one representing the stack of partial syntactic trees, and one representing the history of parse actions to encode parser states (\S\ref{sec:parser}). Since the stack of partial syntactic trees may contain both individual tokens and partial syntactic structures, representations of individual tree fragments are computed compositionally with recursive (i.e., similar to Socher et al., 2014) neural networks.  The parameters are learned with backpropagation (\S\ref{sec:training}), and we obtain state-of-the-art results on Chinese and English dependency parsing tasks (\S\ref{sec:experiments}).

\section{Stack LSTMs}
\label{sec:stacklstms}
In this section we provide a brief review of LSTMs (\S\ref{subsec:lstms}) and then define stack LSTMs (\S\ref{subsec:stacklstms}).

\paragraph{Notation.} We follow the convention that vectors are written with lowercase, boldface letters (e.g., $\mathbf{v}$ or $\mathbf{v}_w$); matrices are written with uppercase, boldface letters (e.g., $\mathbf{M}$, $\mathbf{M}_{a}$, or $\mathbf{M}_{ab}$), and scalars are written as lowercase letters (e.g., $s$ or $q_z$). Structured objects such as sequences of discrete symbols are written with lowercase, bold, italic letters (e.g., $\boldsymbol{w}$ refers to a sequence of input words). Discussion of dimensionality is deferred to the experiments section below (\S\ref{sec:experiments}).

\subsection{Long Short-Term Memories}
\label{subsec:lstms}
LSTMs are a variant of recurrent neural networks (RNNs) designed to cope with the vanishing gradient problem inherent in RNNs \cite{hochreiter:1997,graves:2013}. RNNs read a vector $\mathbf{x}_t$ at each time step and compute a new (hidden) state $\mathbf{h}_t$ by applying a linear map to the concatenation of the previous time step's state $\mathbf{h}_{t-1}$ and the input, and passing this through a logistic sigmoid nonlinearity. Although RNNs can, in principle, model long-range dependencies, training them is difficult in practice since the repeated application of a squashing nonlinearity at each step results in an exponential decay in the error signal through time. LSTMs address this with an extra memory ``cell'' ($\mathbf{c}_t$) that is constructed as a linear combination of the previous state and signal from the input.

LSTM cells process inputs with three multiplicative gates which control what proportion of the current input to pass into the memory cell ($\mathbf{i}_t$) and what proportion of the previous memory cell to ``forget'' ($\mathbf{f}_t$). The updated value of the memory cell after an input $\mathbf{x}_t$ is computed as follows:
\begin{align*}
\mathbf{i}_t &= \sigma(\mathbf{W}_{ix}\mathbf{x}_t + \mathbf{W}_{ih}\mathbf{h}_{t-1} + \mathbf{W}_{ic}\mathbf{c}_{t-1} + \mathbf{b}_i) \\
\mathbf{f}_t &= \sigma(\mathbf{W}_{fx}\mathbf{x}_t + \mathbf{W}_{fh}\mathbf{h}_{t-1} + \mathbf{W}_{fc}\mathbf{c}_{t-1} + \mathbf{b}_f) \\
\mathbf{c}_t &= \mathbf{f}_t \odot \mathbf{c}_{t-1} + \\
& {}\ \ \qquad \mathbf{i}_t \odot \tanh(\mathbf{W}_{cx}\mathbf{x}_t +  \mathbf{W}_{ch}\mathbf{h}_{t-1} + \mathbf{b}_c),
\end{align*}
where $\sigma$ is the component-wise logistic sigmoid function, and $\odot$ is the component-wise (Hadamard) product.

The value $\mathbf{h}_t$ of the LSTM at each time step is controlled by a third gate ($\mathbf{o}_t$) that is applied to the result of the application of a nonlinearity to the memory cell contents:
\begin{align*}
\mathbf{o}_t &= \sigma(\mathbf{W}_{ox}\mathbf{x}_t + \mathbf{W}_{oh}\mathbf{h}_{t-1} + \mathbf{W}_{oc}\mathbf{c}_{t} + \mathbf{b}_o) \\
\mathbf{h}_t &= \mathbf{o}_t \odot \tanh(\mathbf{c}_t).
\end{align*}

To improve the representational capacity of LSTMs (and RNNs generally), LSTMs can be stacked in ``layers'' \cite{pascanu:2014}. In these architectures, the input LSTM at higher layers at time $t$ is the value of $\mathbf{h}_t$ computed by the lower layer (and $\mathbf{x}_t$ is the input at the lowest layer).

Finally, output is produced at each time step from the $\mathbf{h}_t$ value at the top layer:
\begin{align*}
\mathbf{y}_t &= g(\mathbf{h}_t),
\end{align*}
where $g$ is an arbitrary differentiable function.

\subsection{Stack Long Short-Term Memories}
\label{subsec:stacklstms}
Conventional LSTMs model sequences in a left-to-right order.\footnote{Ours is not the first deviation from a strict left-to-right order: previous variations include bidirectional LSTMs \cite{graves:2005} and multidimensional LSTMs \cite{graves:2007}.} Our innovation here is to augment the LSTM with a ``stack pointer.'' Like a conventional LSTM, new inputs are always added in the right-most position, but in stack LSTMs, the current location of the stack pointer determines which cell in the LSTM provides $\mathbf{c}_{t-1}$ and $\mathbf{h}_{t-1}$ when computing the new memory cell contents.

\begin{figure*}
\begin{center}
\includegraphics[scale=0.75]{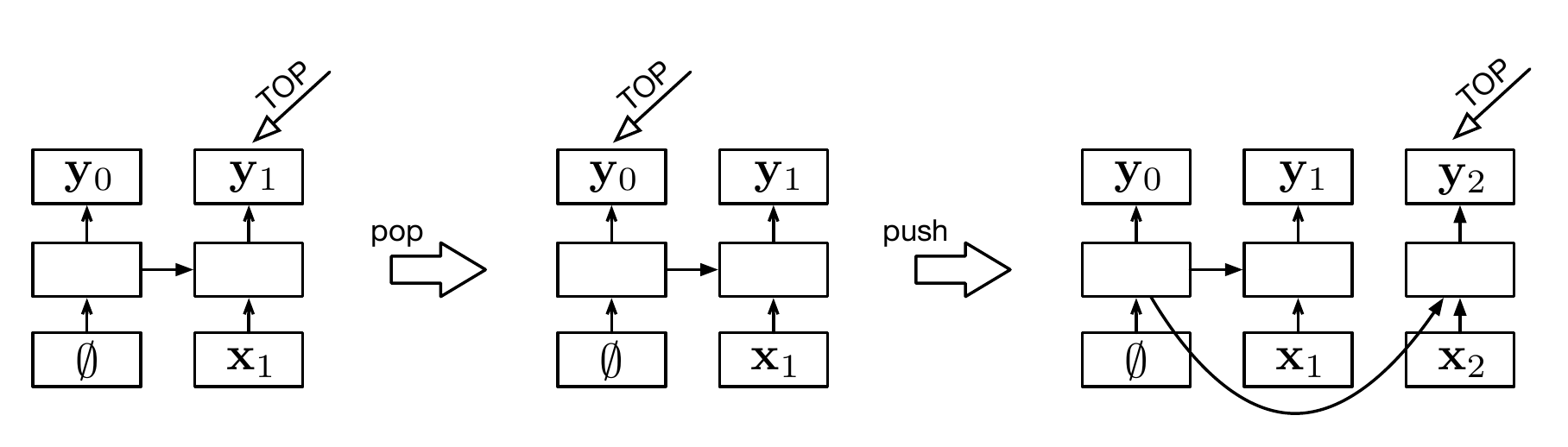}
\vspace{-0.5cm}
\end{center}
\caption{A stack LSTM extends a conventional left-to-right LSTM with the addition of a stack pointer (notated as {\small \textsf{TOP}} in the figure). This figure shows three configurations: a stack with a single element (left), the result of a \textsf{pop} operation to this (middle), and then the result of applying a \textsf{push} operation (right). The boxes in the lowest rows represent stack contents, which are the inputs to the LSTM, the upper rows are the outputs of the LSTM (in this paper, only the output pointed to  by {\small \textsf{TOP}} is ever accessed), and the middle rows are the memory cells (the $\mathbf{c}_t$'s and $\mathbf{h}_t$'s) and gates. Arrows represent function applications (usually affine transformations followed by a nonlinearity), refer to \S\ref{subsec:lstms} for specifics.}
\label{fig:stack_lstm}
\end{figure*}

In addition to adding elements to the end of the sequence, the stack LSTM provides a \textsf{pop} operation which moves the stack pointer to the previous element (i.e., the previous element that was extended, not necessarily the right-most element). Thus, the LSTM can be understood as a stack implemented so that contents are never overwritten, that is, \textsf{push} always adds a new entry at the end of the list that contains a back-pointer to the previous top, and \textsf{pop} only updates the stack pointer.\footnote{\newcite{goldberg:2013} propose a similar stack construction to prevent stack operations from invalidating existing references to the stack in a beam-search parser that must (efficiently) maintain a priority queue of stacks.} This control structure is schematized in Figure~\ref{fig:stack_lstm}.

By querying the output vector to which the stack pointer points (i.e., the $\mathbf{h}_{\textsf{TOP}}$), a continuous-space ``summary'' of the contents of the current stack configuration is available. We refer to this value as the ``stack summary.''\ignore{One might ask why we go through this extra effort to preserve value that would be ``erased'' by a pop operation. The reason is that although these values do not determine the future behavior of the stack (or its summaries), they are necessary for backpropagation of an error signal that is dependent on these elements. \nascomment{this is something I did not understand originally.  It might be helpful to clarify/elaborate.}}

\paragraph{What does the stack summary look like?} Intuitively, elements near the top of the stack will influence the representation of the stack. However, the LSTM has the flexibility to learn to extract information from arbitrary points in the stack \cite{hochreiter:1997}.

Although this architecture is to the best of our knowledge novel, it is reminiscent of the Recurrent Neural Network Pushdown Automaton~(NNPDA) of \newcite{das:1992}, which added an external stack memory to an RNN. However, our architecture provides an embedding of the complete contents of the stack, whereas theirs made only the top of the stack visible to the RNN.

\section{Dependency Parser}
\label{sec:parser}
We now turn to the problem of learning representations of dependency parsers. We preserve the standard data structures of a transition-based dependency parser, namely a buffer of words ($B$) to be processed and a stack ($S$) of partially constructed syntactic elements. Each stack element is augmented with a continuous-space vector embedding representing a word and, in the case of $S$, any of its syntactic dependents. Additionally, we introduce a third stack ($A$) to represent the history of actions taken by the parser.\footnote{The $A$ stack is only ever pushed to; our use of a stack here is purely for implementational and expository convenience.} Each of these stacks is associated with a stack LSTM that provides an encoding of their current contents. The full architecture is illustrated in Figure~\ref{fig:parser}, and we will review each of the components in turn.

\begin{figure*}
\centering
\includegraphics[scale=0.75]{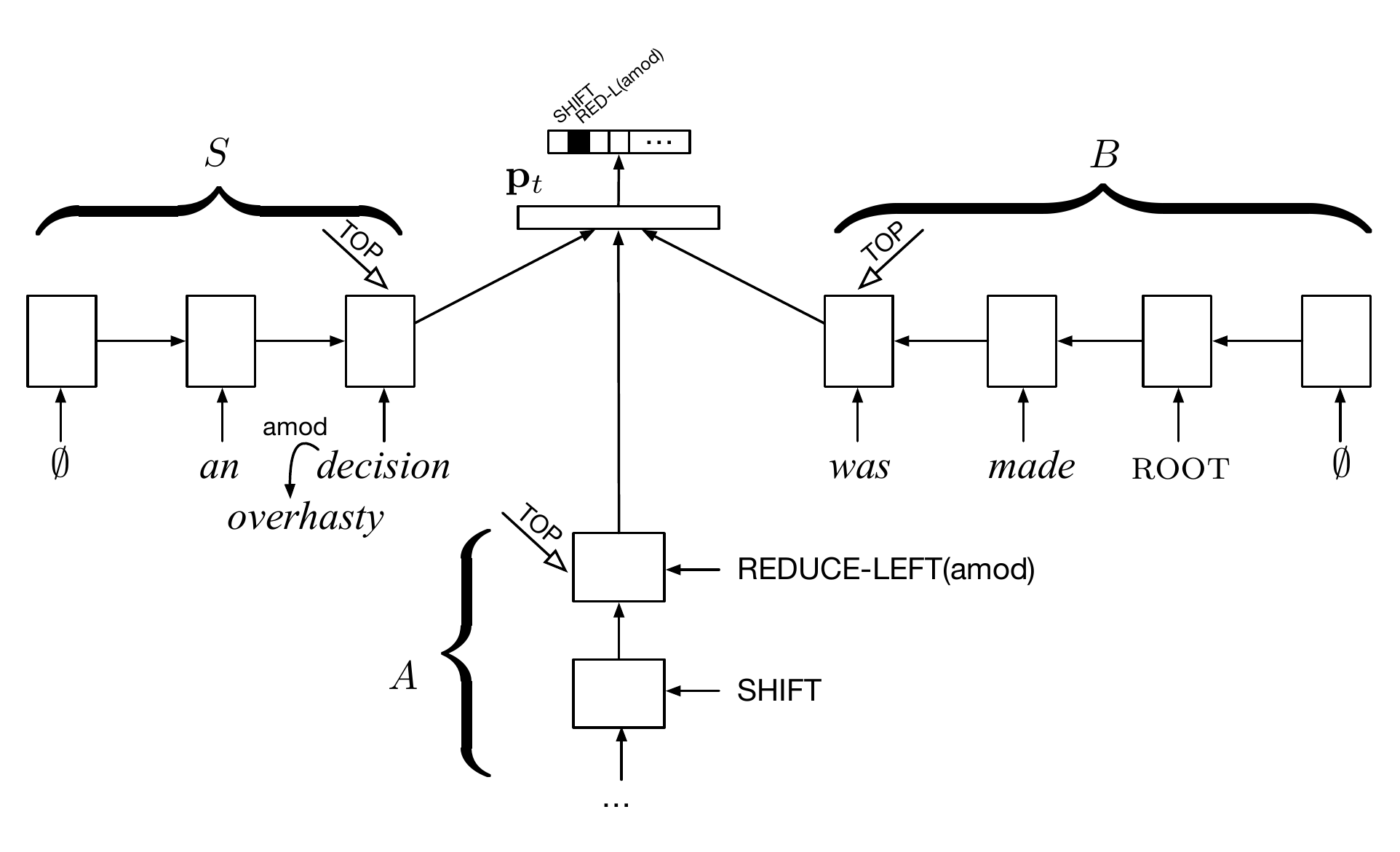}
\vspace{-0.7cm}
\caption{Parser state computation encountered while parsing the sentence ``\emph{an overhasty decision was made}.'' Here $S$ designates the stack of partially constructed dependency subtrees and its LSTM encoding; $B$ is the buffer of words remaining to be processed and its LSTM encoding; and $A$ is the stack representing the history of actions taken by the parser. These are linearly transformed, passed through a ReLU nonlinearity to produce the parser state embedding $\mathbf{p}_t$. An affine transformation of this embedding is passed to a softmax layer to give a distribution over parsing decisions that can be taken.}
\label{fig:parser}
\end{figure*}

\subsection{Parser Operation}
The dependency parser is initialized by pushing the words and their representations (we discuss word representations below in \S\ref{sec:words}) of the input sentence in reverse order onto $B$ such that the first word is at the top of $B$ and the \textsc{root} symbol is at the bottom, and $S$ and $A$ each contain an empty-stack token. At each time step, the parser computes a composite representation of the stack states (as determined by the current configurations of $B$, $S$, and $A$) and uses that to predict an action to take, which updates the stacks. Processing completes when $B$ is empty (except for the empty-stack symbol), $S$ contains two elements, one representing the full parse tree headed by the \textsc{root} symbol and the other the empty-stack symbol, and $A$ is the history of operations taken by the parser.

The parser state representation at time $t$, which we write $\mathbf{p}_t$, which is used to is determine the transition to take, is defined as follows:
\begin{align*}
\mathbf{p}_t = \max \left\{\mathbf{0}, \mathbf{W}[\mathbf{s}_t; \mathbf{b}_t; \mathbf{a}_t] + \mathbf{d}\right\},
\end{align*}
where $\mathbf{W}$ is a learned parameter matrix, $\mathbf{b}_t$ is the stack LSTM encoding of the input buffer $B$, $\mathbf{s}_t$ is the stack LSTM encoding of $S$, $\mathbf{a}_t$ is the stack LSTM encoding of $A$, $\mathbf{d}$ is a bias term, then passed through a component-wise rectified linear unit (ReLU) nonlinearity \cite{glorot:2011}.\footnote{In preliminary experiments, we tried several nonlinearities and found ReLU to work slightly better than the others.}

Finally, the parser state $\mathbf{p}_t$ is used to compute the probability of the parser action at time $t$ as:
\begin{align*}
p(z_t \mid \mathbf{p}_t) = \frac{\exp \left( \mathbf{g}_{z_t}^{\top} \mathbf{p}_t + q_{z_t} \right)}{\sum_{z' \in \mathcal{A}(S,B)} \exp \left( \mathbf{g}_{z'}^{\top} \mathbf{p}_t + q_{z'} \right)} ,
\end{align*}
where $\mathbf{g}_z$ is a column vector representing the (output) embedding of the parser action $z$, and $q_z$ is a bias term for action $z$. The set $\mathcal{A}(S,B)$ represents the valid actions that may be taken given the current contents of the stack and buffer.\footnote{In general, $\mathcal{A}(S,B)$ is the complete set of parser actions discussed in \S\ref{sec:transitions}, but in some cases not all actions are available. For example, when $S$ is empty and words remain in $B$, a \textsc{shift} operation is obligatory \cite{sartorio-satta-nivre:2013:ACL2013}. \ignore{\cjd{Miguel, is there a good reference for these constraints?}}} Since $\mathbf{p}_t = f(\mathbf{s}_t,\mathbf{b}_t,\mathbf{a}_t)$ encodes information about all previous decisions made by the parser, the chain rule may be invoked to write the probability of any valid sequence of parse actions $\boldsymbol{z}$ conditional on the input as:
\begin{align}
p(\boldsymbol{z} \mid \boldsymbol{w}) = \prod_{t=1}^{|\boldsymbol{z}|} p(z_t \mid \mathbf{p}_t). \label{eq:objective}
\end{align}
\subsection{Transition Operations}
\label{sec:transitions}
Our parser is based on the arc-standard transition inventory \cite{nivre2004}, given in Figure~\ref{fig:parser}.

\begin{figure*}
\centering
\begin{tabular}{cc|l|cc|c}
\textbf{Stack}$_t$ & \textbf{Buffer}$_t$ & \textbf{Action} & \textbf{Stack}$_{t+1}$ & \textbf{Buffer}$_{t+1}$ & \textbf{Dependency} \\
\hline
$(\mathbf{u},u),(\mathbf{v},v),S$ & $B$  &$\textsc{reduce-right}(r)$ & $(g_r(\mathbf{u},\mathbf{v}),u),S$ & $B$ & $u \stackrel{\scriptsize{r}}{\rightarrow} v$ \\
$(\mathbf{u},u),(\mathbf{v},v), S$ & $B$ & $\textsc{reduce-left}(r)$ & $(g_r(\mathbf{v},\mathbf{u}),v), S$ & $B$ & $u \stackrel{\scriptsize{r}}{\leftarrow} v$ \\
$S$ & $(\mathbf{u},u),B$ & \textsc{shift} & $(\mathbf{u},u),S$ & $B$ & --- 
\end{tabular}
\caption{\label{fig:parser}Parser transitions indicating the action applied to the stack and buffer and the resulting stack and buffer states. Bold symbols indicate (learned) embeddings of words and relations, script symbols indicate the corresponding words and relations.}
\end{figure*}

\paragraph{Why arc-standard?} 
Arc-standard transitions parse a sentence from left to right, using a stack to store partially built syntactic structures and a buffer that keeps the incoming tokens to be parsed. The parsing algorithm chooses an action at each configuration by means of a score. In arc-standard parsing, the dependency tree is constructed bottom-up, because right-dependents of a head are only attached after the subtree under the dependent is fully parsed. Since our parser recursively computes representations of tree fragments, this construction order guarantees that once a syntactic structure has been used to modify a head, the algorithm will not try to find another head for the dependent structure. This means we can evaluate composed representations of tree fragments incrementally; we discuss our strategy for this below (\S\ref{sec:comp}).

\ignore{
 simplifies the complexity of evaluating the recursively composed representation of a tree fragments. Furthermore, since our parser has the ability to look ``deeply'' into the buffer and stack (and its constructed items), in principle, it avoids the difficulties of architecting features. \nascomment{last two sentences are inside baseball.  you're assuming the reader has some other thing in mind, but you're not saying what that thing is.  simpliefies complexity relative to what?  what is the implied thing that has ``difficulties of architecting features''?}}

\subsection{Token Embeddings and OOVs}
\label{sec:words}

\begin{figure}[h]
\hspace{-.25cm}\includegraphics[scale=0.68]{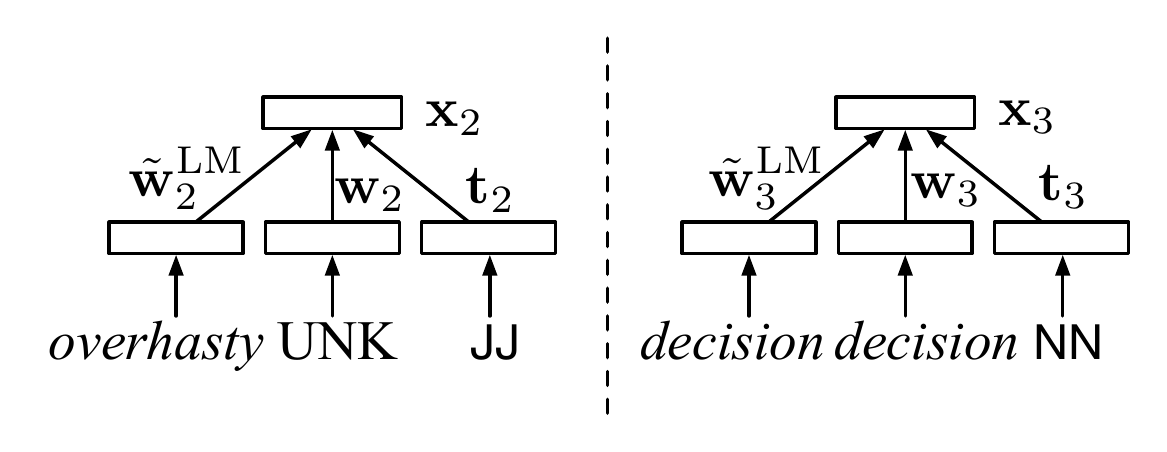}
\vspace{-0.8cm}
\caption{Token embedding of the words \emph{decision}, which is present in both the parser's training data and the language model data, and \emph{overhasty}, an adjective that is not present in the parser's training data but is present in the LM data.
\label{fig:tokens}}
\end{figure}

To represent each input token, we concatenate three vectors: a learned vector representation for each word type ($\mathbf{w}$);
a fixed vector representation from a neural language model ($\tilde{\mathbf{w}}_{\textrm{LM}}$), and a learned representation ($\mathbf{t}$) of the POS tag of the token, provided as auxiliary input to the parser. A linear map ($\mathbf{V}$) is applied to the resulting vector and passed through a component-wise ReLU,
\begin{align*}
\mathbf{x} = \max\left\{\mathbf{0}, \mathbf{V}[\mathbf{w}; \tilde{\mathbf{w}}_{\textrm{LM}}; \mathbf{t}]  + \mathbf{b} \right\} .
\end{align*}
This mapping can be shown schematically as in Figure~\ref{fig:tokens}.

This architecture lets us deal flexibly with out-of-vocabulary words---both those that are OOV in both the very limited parsing data but present in the pretraining LM, and words that are OOV in both. To ensure we have estimates of the OOVs in the parsing training data, we stochastically replace (with $p=0.5$)  each singleton word type in the parsing training data with the UNK token in each training iteration.

\paragraph{Pretrained word embeddings.} A veritable cottage industry exists for creating word embeddings, meaning numerous pretraining options for $\mathbf{\tilde{w}}_{\textrm{LM}}$ are available. However, for syntax modeling problems, embedding approaches which discard order perform less well \cite{bansal2014tailoring}; therefore we used a variant of the skip $n$-gram model introduced by \newcite{Ling:2015:naacl}, named ``structured skip $n$-gram,'' where a different set of parameters is used to predict each context word depending on its position relative to the target word. The hyperparameters of the model are the same as in the skip $n$-gram model defined in word2vec~\cite{mikolov2013distributed}, and we set the window size to 5, used a negative sampling rate to 10, and ran 5 epochs through unannotated corpora described in \S\ref{sec:data}.

\subsection{Composition Functions}
\label{sec:comp}
Recursive neural network models enable complex phrases to be represented compositionally in terms of their parts and the relations that link them \cite{socher:2011,socher:2013,hermann:2013,socher:2013b}. We follow this previous line of work in embedding dependency tree fragments that are present in the stack $S$ in the same vector space as the token embeddings discussed above.

\begin{figure}[h]
\begin{center}
\includegraphics[scale=0.75]{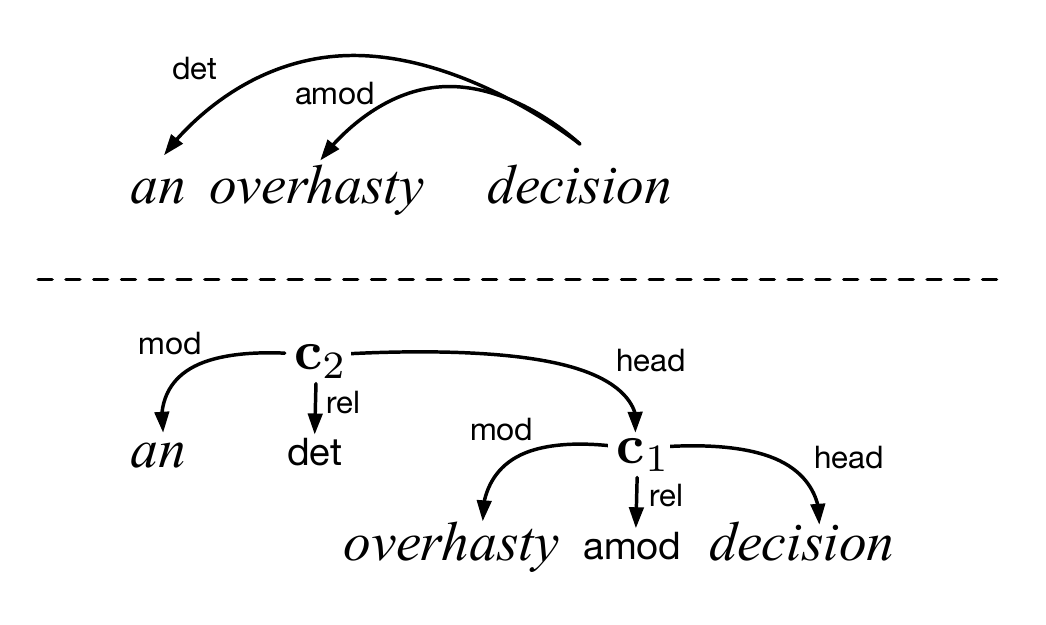}
\vspace{-0.7cm}
\end{center}
\caption{The representation of a dependency subtree (above) is computed by recursively applying composition functions to $\langle \textrm{head}, \textrm{modifier}, \textrm{relation} \rangle$ triples. In the case of multiple dependents of a single head, the recursive branching order is imposed by the order of the parser's reduce operations (below).}
\label{fig:composition}
\end{figure}

A particular challenge here is that a syntactic head may, in general, have an arbitrary number of dependents. To simplify the parameterization of our composition function, we combine head-modifier pairs one at a time, building up more complicated structures in the order they are ``reduced'' in the parser, as illustrated in Figure~\ref{fig:composition}. Each node in this expanded syntactic tree has a value computed as a function of its three arguments: the syntactic head ($\mathbf{h}$), the dependent ($\mathbf{d}$), and the syntactic relation being satisfied ($\mathbf{r}$). We define this  by concatenating the vector embeddings of the head, dependent and relation, applying a linear operator and a component-wise nonlinearity as follows:
\begin{align*}
\mathbf{c} = \tanh\left(\mathbf{U}[\mathbf{h}; \mathbf{d}; \mathbf{r}] + \mathbf{e} \right).
\end{align*}
For the relation vector, we use an embedding of the parser action that was applied to construct the relation (i.e., the syntactic relation paired with the direction of attachment).

\section{Training Procedure}
\label{sec:training}
We trained our parser to maximize the conditional log-likelihood (Eq.~\ref{eq:objective}) of treebank parses given sentences. Our implementation constructs a computation graph for each sentence and runs forward- and backpropagation to obtain the gradients of this objective with respect to the model parameters. The computations for a single parsing model were run on a single thread on a CPU. Using the dimensions discussed in the next section, we required between 8 and 12 hours to reach convergence on a held-out dev set.\footnote{Software for replicating the experiments is available from \url{https://github.com/clab/lstm-parser}.}

Parameter optimization was performed using stochastic gradient descent with an initial learning rate of $\eta_0=0.1$, and the learning rate was updated on each pass through the training data as $\eta_t = \eta_0/(1+\rho t)$, with $\rho=0.1$ and where $t$ is the number of epochs completed. No momentum was used. To mitigate the effects of ``exploding'' gradients, we clipped the $\ell_2$ norm of the gradient to 5 before applying the weight  update rule \cite{sutskever:2014,graves:2013}. An $\ell_2$ penalty of $1 \times 10^{-6}$ was applied to all weights.

Matrix and vector parameters were initialized with uniform samples in $\pm \sqrt{6/(r+c)}$, where $r$ and $c$ were the number of rows and columns in the structure \cite{glorot:2010}.


\paragraph{Dimensionality.}
The full version of our parsing model sets dimensionalities as follows.  LSTM hidden states are of size 100, and we use two layers of LSTMs for each stack. Embeddings of the parser actions used in the composition functions have 16 dimensions, and the output embedding size is 20 dimensions. Pretained word embeddings have 100 dimensions (English) and 80 dimensions (Chinese), and the learned word embeddings have 32 dimensions. Part of speech embeddings have 12 dimensions.

These dimensions were chosen based on intuitively reasonable values (words should have higher dimensionality than parsing actions, POS tags, and relations; LSTM states should be relatively large), and it was confirmed on development data that they performed well.\footnote{We did perform preliminary experiments with LSTM states of 32, 50, and 80, but the other dimensions were our initial guesses.}
Future work might more carefully optimize these parameters; our reported architecture strikes a balance between minimizing computational expense and finding solutions that work.

\section{Experiments}
\label{sec:experiments}
We applied our parsing model and several variations of it to two parsing tasks and report results below.

\subsection{Data} \label{sec:data}
We used the same data setup as \newcite{chen:2014}, namely an English and a Chinese parsing task. This baseline configuration was chosen since they likewise used a neural parameterization to predict actions in an arc-standard transition-based parser.


\begin{itemize}
\item For English, we used the Stanford Dependencency (SD) treebank \cite{Marneffe06generatingtyped} used in \cite{chen:2014} which is the closest model published, with the same splits.\footnote{Training: 02-21. Development: 22. Test: 23.} The part-of-speech tags are predicted by using the Stanford Tagger \cite{Toutanova:2003:FPT:1073445.1073478} with an accuracy of 97.3\%. This treebank contains a negligible amount of non-projective arcs \cite{chen:2014}.
 \item For Chinese, we use the Penn Chinese Treebank 5.1 (CTB5) following Zhang and Clark \shortcite{zhang08},\footnote{Training: 001--815, 1001--1136. Development: 886--931, 1148--1151. Test: 816--885, 1137--1147.} with gold part-of-speech tags which is also the same as in \newcite{chen:2014}. 
\ignore{ \item For English, we use the WSJ section of the Penn Treebank, converted with the head-finding rules of Yamada and Matsumoto \shortcite{yamada03} (in the following YM) and the labeling rules of Nivre \shortcite{nivre06book}.\footnote{Training: 02-21. Development: 24. Test: 23.}. We used predicted part-of-speech tags by using the MarMoT tagger \cite{mueller2013} with an accuracy of 97.3\%, assigned in a ten-way jackknifing to the training data. This treebank is fully projective.}
\end{itemize}
Language model word embeddings were generated, for English, from the AFP portion of the English Gigaword corpus (version~5), and from the complete Chinese Gigaword corpus (version~2), as segmented by the Stanford Chinese Segmenter \cite{tseng:2005}.




\subsection{Experimental configurations}
We report results on five experimental configurations per language, as well as the \newcite{chen:2014} baseline. These are: the full stack LSTM parsing model (S-LSTM), the stack LSTM parsing model without POS tags ($-$POS), the stack LSTM parsing model without pretrained language model embeddings ($-$pretraining), the stack LSTM parsing model that uses just head words on the stack instead of composed representations ($-$composition), and the full parsing model where rather than an LSTM, a classical recurrent neural network is used (S-RNN).

\subsection{Results}


Following \newcite{chen:2014} we exclude punctuation symbols for evaluation. Tables \ref{sd-res} and \ref{ctbgold-res} show comparable results with \newcite{chen:2014}, and we show that our model is better than their model in both the development set and the test set.\ignore{\nascomment{well, not really comparable as they use gold POS!  this should be pointed out before now, and ideally we'd also show how we do with gold POS, for a more direct comparison with them.  I'd add that below the double lines in the two tables.}}

\begin{table}[h]
\centering
\begin{tabular}{l|cc|cc}
& \multicolumn{2}{|c|}{Development}&\multicolumn{2}{|c}{Test} \\
 & UAS & LAS & UAS & LAS \\
 \hline
S-LSTM & \textbf{93.2} & \textbf{90.9} & \textbf{93.1} & \textbf{90.9} \\
$-$POS & 93.1 & 90.4 & 92.7 & 90.3 \\
$-$pretraining & 92.7 & 90.4 & 92.4 & 90.0 \\
$-$composition & 92.7 & 89.9 & 92.2 & 89.6 \\
S-RNN & 92.8 & 90.4 & 92.3 & 90.1 \\
\hline
\hline
C\&M (2014) & 92.2 & 89.7 & 91.8 & 89.6
\end{tabular}
\caption{English parsing results (SD)}
\label{sd-res}
\end{table}%

\ignore{
\begin{table}[h]
\centering
\begin{tabular}{l|cc|cc}
 & UAS & LAS & UAS & LAS \\
 \hline
S-LSTM ($b=5$) &  & & &\\
$-$beam & & & & \\
$-$pretraining & & & & \\
$-$POS & & & & \\
S-RNN & & & & \\
\hline
\hline
C\&M (2014) & 92.2 & 91.0 & 92.0 & 90.7
\end{tabular}
\caption{English parsing results (LTH). POS tag accuracy:~97.3\%}
\label{lth-res}
\end{table}
}

\ignore{
\begin{table}[h]
\footnotesize
\centering
\begin{tabular}{l|cc|cc}
 & UAS & LAS & UAS & LAS \\
 \hline
S-LSTM ($b=5$) &  & & & \\
$-$beam & & &  \\
$-$pretraining & & & & \\
$-$POS & & & & \\
S-RNN & & & &\\
\hline
\hline
\newcite{mcdonald05acl}               &90.9~~ &   \\ 
\newcite{mcdonald06eacl}              &91.5~~ &   \\
\newcite{huang10}                     &92.1~~ &   \\
\newcite{koo10acl}                    &93.04&     \\ 
\newcite{zhang-nivre:2011:ACL-HLT2011}&92.9~~ &   \\  
\newcite{ballesteros2014automatic}  & 93.49  & 92.53 & -- \\ 
\hline
\newcite{koo08} $\dagger$             &93.16 & \\
\newcite{carreras08} $\dagger$        &93.5~~ &  \\ 
\newcite{suzuki09}  $\dagger$         &93.79&       \\ \hline 
\end{tabular}
\caption{English parsing results (YM) with state-of-the-art comparison with predicted POS tags. POS tag accuracy: ~97.3\%. Results marked with $\dagger$ use additional information sources and are not directly comparable to the others. }
\label{ym-res}
\end{table}
}
\begin{table}[h]
\centering
\begin{tabular}{l|cc|cc}
& \multicolumn{2}{|c|}{Dev. set}&\multicolumn{2}{|c}{Test set} \\
\hline
 & UAS & LAS & UAS & LAS \\
 \hline
S-LSTM & \textbf{87.2} & \textbf{85.9} & \textbf{87.2} & \textbf{85.7} \\
$-$composition & 85.8 & 84.0 & 85.3 & 83.6 \\
$-$pretraining & 86.3 & 84.7 & 85.7 & 84.1 \\
$-$POS & 82.8 & 79.8 & 82.2 & 79.1 \\
S-RNN & 86.3 & 84.7 & 86.1 & 84.6 \\
\hline
\hline
C\&M (2014) & 84.0 & 82.4 & 83.9 & 82.4
\end{tabular}
\caption{Chinese parsing results (CTB5)}
\label{ctbgold-res}
\end{table}%

\ignore{
\begin{table}[h]
\footnotesize
\centering
\begin{tabular}{l|cc|cc}
 & UAS & LAS & UAS & LAS \\
 \hline
S-LSTM & 87.2 & & &\\
$-$beam & 84.9 & & & \\
$-$composition & & & & \\
$-$pretraining & & & & \\
$-$POS & & & & \\
S-RNN & & & & \\
\hline
\hline
\newcite{li11} 3rd-order                       & 80.60  &&\\ 
\newcite{hatori11} HS                          & 79.60  &&\\
\newcite{hatori11} ZN                          & 81.20  &&\\
\newcite{ballesteros2014automatic}           & 81.77  & 94.28 \\
\hline
\end{tabular}
\caption{Chinese parsing results with predicted POS tags (CTB5). POS tag accuracy:~93.9\%.}
\label{ctb-res}
\end{table}
}
\subsection{Analysis}
Overall, our parser substantially outperforms the baseline neural network parser of \newcite{chen:2014}, both in the full configuration and in the various ablated conditions we report. The one exception to this is the $-$POS condition for the Chinese parsing task, which in which we underperform their baseline (which used gold POS tags), although we do still obtain reasonable parsing performance in this limited case. We note that predicted POS tags in English add very little value---suggesting that we can think of parsing sentences directly without first tagging them. We also find that using composed representations of dependency tree fragments outperforms using representations of head words alone, which has implications for theories of headedness. Finally, we find that while LSTMs outperform baselines that use only classical RNNs, these are still quite capable of learning good representations.

\paragraph{Effect of beam size.} Beam search was determined to have minimal impact on scores (absolute improvements of $\le 0.3\%$ were possible with small beams). Therefore, all results we report used greedy decoding---\newcite{chen:2014} likewise only report results with greedy decoding. This finding is in line with previous work that generates sequences from recurrent networks \cite{grefenstette:2014}, although \newcite{vinyals:2015} did report much more substantial improvements with beam search on their ``grammar as a foreign language'' parser.\footnote{Although superficially similar to ours, \newcite{vinyals:2015} is a phrase-structure parser and adaptation to the dependency parsing scenario would have been nontrivial. We discuss their work in \S\ref{sec:related}.}

\ignore{
\subsection{Analysis}
What can the parameters tell us about syntax?
As discussed above, we reasoned that left- and right-attachments behave differently with respect to the syntactic properties of the heads they project, so we learned independent relation embeddings for each (direction, relation) type used in the composition function. Figure~\ref{fig:learned-rels} plots a projection of these vectors, allowing us to see to what extent this assumption is true. Briefly, we see that while \textsc{amod} and \textsc{nmod} relations have similar representations, independent of the direction they modify, other relations, such as \textsc{vmod} and  \textsc{sub} do indeed have different representations. Furthermore, it is interesting to observe that \textsc{r(obj)} and \textsc{r(vmod)}---the relations used to link post-verbal arguments (respectively, adjuncts) to their governing verb---are quite close. This arguably captures the intuitive ambiguity that exists for speakers, linguists, and parsers about the dividing line between arguments and adjuncts.

\begin{figure}[h]
\hspace{-.5cm}\includegraphics[scale=0.47]{english-rels.pdf}
\vspace{-1.5cm}
\caption{Learned action embeddings used as arguments to recursive composition function (English).}
\label{fig:learned-rels}
\end{figure}
}

\section{Related Work}
\label{sec:related}

Our approach ties together several strands of previous work. First, several kinds of stack memories have been proposed to augment neural architectures. \newcite{das:1992} proposed a neural network with an external stack memory based on recurrent neural networks. In contrast to our model, in which the entire contents of the stack are summarized in a single value, in their model, the network could only see the contents of the top of the stack. \newcite{miikkulainen:1996} proposed an architecture with a stack that had a summary feature, although the stack control was learned as a latent variable.

A variety of authors have used neural networks to predict parser actions in shift-reduce parsers. The earliest attempt we are aware of is due to \newcite{mayberry:1999}. The resurgence of interest in neural networks has resulted in in several applications to transition-based dependency parsers \cite{weiss:2015,chen:2014,stenetorp:2013}.  In these works, the conditioning structure was manually crafted and sensitive to only certain properties of the state, while we are conditioning on the global state object.  Like us, \newcite{stenetorp:2013} used recursively composed representations of the tree fragments (a head and its dependents). Neural networks have also been used to learn representations for use in chart parsing \cite{henderson:2004,titov:2007,socher:2013c,le:2014}.

LSTMs have also recently been demonstrated as a mechanism for learning to represent parse structure.\newcite{vinyals:2015} proposed a phrase-structure parser based on LSTMs which operated by first reading the entire input sentence in so as to obtain a vector representation of it, and then generating bracketing structures sequentially conditioned on this representation. Although superficially similar to our model, their approach has a number of disadvantages. First, they relied on a large amount of semi-supervised training data that was generated by parsing a large unannotated corpus with an off-the-shelf parser. Second, while they recognized that a stack-like shift-reduce parser control provided useful information, they only made the top word of the stack visible during training and decoding. Third, although it is impressive feat of learning that an entire parse tree be represented by a vector, it seems that this formulation makes the problem unnecessarily difficult.

Finally, our work can be understood as a progression toward using larger contexts in parsing. An exhaustive summary is beyond the scope of this paper, but some of the important milestones in this tradition are the use of cube pruning to efficiently include nonlocal features in discriminative chart reranking \cite{huang:2008}, approximate decoding techniques based on LP relaxations in graph-based parsing to include higher-order features \cite{martins:2010}, and randomized hill-climbing methods that enable arbitrary nonlocal features in global discriminative parsing models \cite{zhang:2014}. Since our parser is sensitive to any part of the input, its history, or its stack contents, it is similar in spirit to the last approach, which permits truly arbitrary features.

\section{Conclusion}
We presented stack LSTMs, recurrent neural networks for sequences, with push and pop operations, and used them to implement a state-of-the-art transition-based dependency parser. 
We conclude by remarking that stack memory offers intriguing possibilities for learning to solve general information processing problems \cite{miikkulainen:1996}.  Here, we learned from observable stack manipulation operations (i.e., supervision from a treebank), and the computed embeddings
of final parser states were not used for any further prediction.  However, this could be reversed, giving a device that learns to construct context-free programs (e.g., expression trees) given only observed outputs; one application would be unsupervised parsing.
Such an extension of the work would make it an alternative to architectures that have an explicit external memory such as neural Turing machines \cite{DBLP:journals/corr/GravesWD14} and memory networks \cite{weston:2015}. However, as with those models, without supervision of the stack operations, formidable computational challenges must be solved (e.g., marginalizing over all latent stack operations), but sampling techniques and techniques from reinforcement learning have promise here \cite{zaremba:2015}, making this an intriguing avenue for future work.

\section*{Acknowledgments}
The authors would like to thank Lingpeng Kong and Jacob Eisenstein for comments on an earlier version of this draft and Danqi Chen for assistance with the parsing datasets. This work was sponsored in part by the U.~S.~Army Research Laboratory and the U.~S.~Army Research Office under contract/grant number W911NF-10-1-0533, and in part by NSF CAREER grant IIS-1054319. Miguel Ballesteros is supported by the European Commission under the contract numbers FP7-ICT-610411 (project MULTISENSOR) and H2020-RIA-645012 (project KRISTINA).

\bibliographystyle{acl}
\bibliography{biblio,main}

\end{document}